\UseRawInputEncoding

\documentclass[journal]{IEEEtran}

\usepackage{xcolor,soul,framed} 

\colorlet{shadecolor}{yellow}
\usepackage[pdftex]{graphicx}
\graphicspath{{../pdf/}{../jpeg/}}
\DeclareGraphicsExtensions{.pdf,.jpeg,.png}

\usepackage[cmex10]{amsmath}
\usepackage{array}
\usepackage{mdwmath}
\usepackage{mdwtab}
\usepackage{eqparbox}
\usepackage{url}

\usepackage{comment} 
\usepackage{bm}
\usepackage{mathtools}  
\usepackage{textcomp}   
\usepackage{gensymb}    
\usepackage[shortcuts]{extdash}
\usepackage{hyperref}
\usepackage{amssymb}

\usepackage{xcolor}         

\DeclareMathOperator{\Tr}{Tr}

\begin{document}
\bstctlcite{IEEEexample:BSTcontrol}
    \title{MIRRAX: A Reconfigurable Robot for Limited Access Environments}
  \author{Wei Cheah, Keir Groves, Horatio Martin, Harriet Peel, Simon Watson, Ognjen Marjanovic and Barry Lennox
  

  \thanks{This work was supported by UK Research and Innovation through the Engineering and Physical Science Research Council under grant number EP/P01366X/1 and EP/R026084/1 and the Royal Academy of Engineering under grant number CiET1819/13}
  \thanks{For the purpose of open access, the author has applied a Creative Commons Attribution (CC BY) licence licence may be stated instead) to any Author Accepted Manuscript version arising.}
  \thanks{The authors are all with the Department of Electrical and Electronic Engineering, The University of Manchester, UK. }%
  }  

\markboth{ACCEPTED ON THE IEEE TRANSACTIONS ON ROBOTICS.}{Wei \MakeLowercase{\textit{et al.}}: MIRRAX: A Reconfigurable Robot for Limited Access Environments}

\maketitle

\begin{abstract}
The development of mobile robot platforms for inspection has gained traction in recent years. However, conventional mobile robots are unable to address the challenge of operating in extreme environments where the robot is required to traverse narrow gaps in highly cluttered areas with restricted access, typically through narrow ports. This paper presents MIRRAX, a robot designed to meet these challenges by way of its reconfigurable capability. Controllers for the robot are detailed, along with an analysis on the controllability of the robot given the use of Mecanum wheels in a variable configuration. Characterisation on the robot's performance identified suitable configurations for operating in narrow environments. Experimental validation of the robot's controllability shows good agreement with the theoretical analysis, and the capability to address the challenges of accessing entry ports as small as 150mm diameter, as well as navigating through cluttered environments. The paper also presents results from a deployment in a Magnox facility at the Sellafield nuclear site in the UK -- the first robot to ever do so, for remote inspection and mapping.
\end{abstract}


\IEEEpeerreviewmaketitle

\section{Introduction}\label{s:intro}

\IEEEPARstart{T}{he} need for inspection, maintenance and repair (IMR) in extreme environments has grown considerably following the Fukushima incident and the continued aging of nuclear facilities \cite{zhao2017}. These tasks are non-trivial, requiring substantial cost, and where human operators are involved, there is an element of risk to their health when working in these environments. Faced with these challenges, the use of mobile robots has gained interest as risks to human health can be mitigated and robots can, in some cases, perform tasks that humans are unable to \cite{Tsitsimpelis2019}. Even so, challenges remain in deploying and operating mobile robots in these environments.
One such challenge is navigating through constrained and cluttered areas and another is that access points may be restricted, for example, many areas in legacy nuclear facilities can only be accessed through a 150 \.mm access port~\cite{Tepco2017}.

Re-configurable terrestrial robots with movable joints offer a solution to these challenges. A typical robot setup would comprise two or more rigid bodies that are connected by movable joints, with locomotion provided by wheels or tracks. 
The range of configurations allow such robots to access areas through small ports and to traverse through highly constrained environments such as inside pipes, ducts or between tightly spaced obstacles. Notable platforms that have been developed for such environments typically employ the use of tracks in their driving mechanisms, due to the significant presence of rubble and loose dirt \cite{Tepco2017,Sarcos}. However, for semi-structured environments such as legacy nuclear facilities, alternative driving mechanism can be used since the terrains encountered can be considered more traversable.   

\begin{figure}[htb]
    \centerline{\includegraphics[width=8.9cm]{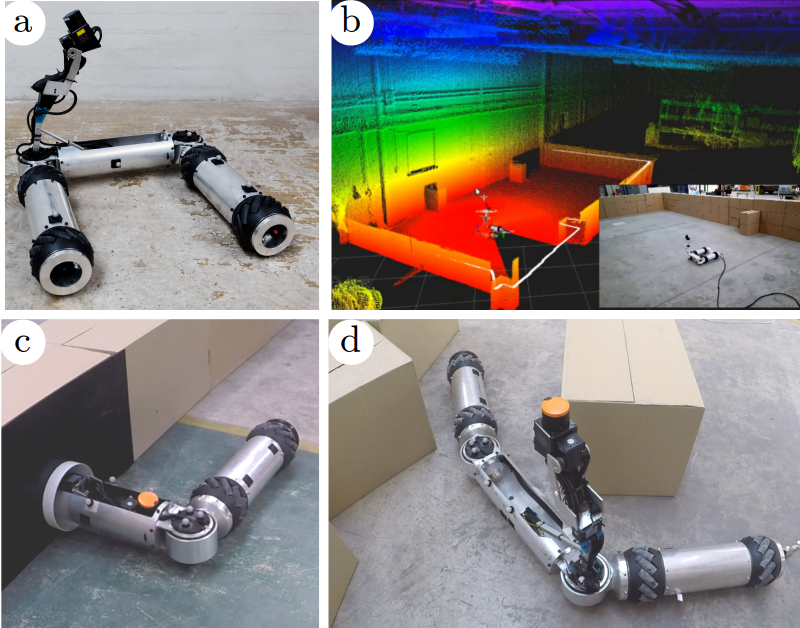}}
    \caption{The MIRRAX re-configurable robot (a)~Default configuration with raised arm for sensor payload (b)~3D point cloud reconstruction (c)~Egress from a 150\,mm access port (d)~Cluttered area navigation.}
    \label{fig:mirrax}
\end{figure}

To this end, we propose the Miniature Inspection Robot for Restricted Access eXploration (MIRRAX), shown in Fig.~\ref{fig:mirrax}a. Sensors mounted on the arm enables 3D reconstruction (see Fig.~\ref{fig:mirrax}b) or other mission-specific tasks.
The design consists of four Mecanum wheels, each pair connected to an actuated link which is in turn connected to the base link. 
The design of the robot enables it to reconfigure and propel itself through a 150\,mm access port, as well as reconfigure itself to navigate through narrow pathways (see Fig.~\ref{fig:mirrax}c\=/d). 
The contribution of this paper are as follows:
\begin{enumerate}
  \item The mechatronic design of a novel reconfigurable mecanum-wheel driven robot,
  \item Controllability analysis for reconfigurable mecanum-wheel driven robot,
  \item Demonstration and analysis on the basic capabilities of the robot, 
  \item Deployment in Sellafield's Magnox facility. 
\end{enumerate}

The remainder of this paper are arranged as follows.
Section~\ref{s:related} reviews related work on reconfigurable robots that has the potential to address the 150~mm access port challenge.
Section~\ref{s:hardware} presents the hardware design of the robot. 
Section~\ref{s:modelling} presents the kinematic and dynamic models used for the controllability proofs in Section~\ref{s:controllability}, as well as the feedback control used for trajectory tracking. 
Section~\ref{s:experiment} demonstrates omni-directional capability (controllability) for non-standard Mecanum wheel configurations through a series of experiments and ingress of 150~mm access port\footnote{A video of the experiments is available in the supplementary material.}, as well as limitations of the robot. 
Section~\ref{s:deploy} details the deployment of the robot in the Magnox facility in Sellafield and lessons learned. 
Finally, Section~\ref{s:conclusion} concludes the work and provides an outlook to future work.

\section{Related Works}\label{s:related}
A comprehensive review of ground mobile robots used in the nuclear industry is presented in \cite{Tsitsimpelis2019}. Many of these environments are highly cluttered due to machinery and there are often areas which cannot be explored using traditional platforms due to their size. This is further compounded by the restriction on the access port for accessing such facilities to 150~mm diameter. 
With these restrictions in place, the choice of using reconfigurable robots, though more complicated, can address these difficult restrictions.

\subsection{Reconfigurable Robots}
In a recent case study~\cite{Sarcos}, Sarcos' Guardian~S re-configurable robot was used to inspect a dust extraction system, reducing both risk to employees and manufacturing downtime. Another example was developed by Tokyo Electric Power Company~(TEPCO)~\cite{Tepco2017}. Their robot was designed to take a tubular form when traversing down the narrow pipe. Once clear of the pipe, the robot expands into a U~shape for stability, similar to fixed-configuration tracked robots, to carry inspection task.

Other forms of mobile robotic platforms that have been used for restricted access environment exploration, though may not specifically fulfil the access port requirement in this study, are modular re-configurable \cite{Seo2019} and snake-type robots \cite{Hirose2009}. Modular re-configurable robots consist of multiple detachable modules that can either be similar in design and capability (homogeneous) \cite{Liu2019,Liang2020} or dissimilar (heterogeneous) \cite{Ahmadzadeh2016}. 
The latter enables a larger variety of payload or functionality to be deployed simultaneously \cite{Pfotzer2014,Romanov2021}. These modules can be combined in a variety of shapes to meet the required demands such as stair-climbing \cite{Liu2005} and stepping over obstacles \cite{Yim2000}.

However reconfigurable robots presents additional challenges in the design and application. The mechanical and control system is evidently more complex compared to a reconfigurable robot. The connection mechanism for mechanical linkage, power and communication between modules needs to be robust and support the required communication bandwidth, especially where high speed communication is required.
In hazardous environment, operating the robot with a tether provide a way of full retrieval of the robot in the event the robot is rendered inoperable due to radiation damage or unforseen circumstances \cite{Bird2021}. Using modular reconfigurable robots introduces the possibility that damaged modules detached from the main body becomes irretrievable. 

The structure of snake-like robot typically consists of a serial chain of non-detachable modules connected  through an actuated joint. 
Motion of such robots depends on the drive mechanism employed. For bio-inspired approaches, each module has at least an actuated rotary motion. Motion is achieved through undulation along the robot's body, enabling motion in both open area \cite{Liljeback2011} and in the presence of obstacles \cite{Trebuna2016,Transeth2008}. 
Alternatively, modules with drive mechanism such as track \cite{Mooney2014} or wheels \cite{Pfotzer2017} enable motion without undulation of the body.

In general, reconfigurable robots capable of fulfilling both the access requirements and have the capability for large sensor payloads are limited in the literature. 
The size of the modules generally limits the available payload size and mass. The actuation mechanism used for realising motion, for example undulation of the robot's body, further restricts feasible mounting locations for sensor payloads.  
Such robots are especially valuable for nuclear environments, as seen by the case study from TEPCO.

\subsection{Controllability of Mecanum Wheels}
\begin{figure}[h]
	\begin{centering}
		\includegraphics[width=9cm]{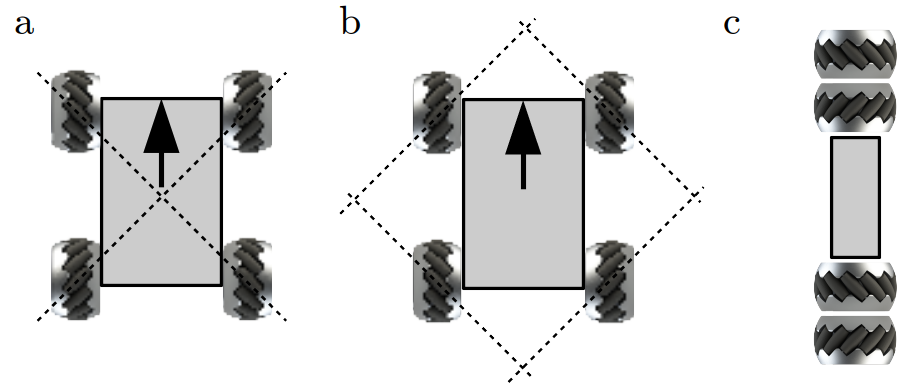}
		\par\end{centering}
	\caption{Mecanum wheel arangements (a) X-configuration (standard) (b) O-configuration (c) Straight line configuration.}
	\label{fig:wheel_arrange}
\end{figure}

The use of Mecanum wheels in a re-configurable manner deviates significantly from the literature and commercial robots. Mecanum wheels are commonly arranged in a rectangular shape, where the rollers form an X-shape when viewed from above (see Fig.~\ref{fig:wheel_arrange}) \cite{Cooney2004,Bischoff2011,Keek2019}. The same arrangement is observed for platforms with more than four wheels \cite{Kukaomnimove2021}. More recently, Mecanum wheels arranged in a straight line have been proposed \cite{Reynolds-haertle2011,Watson2018}. Compared to ballbots \cite{Lauwers2007} which have a single contact point with the ground and are non-holonomic, a straight line configuration with Mecanum wheels has a larger number of contact points and is holonomic.

With regards to the controllability of using Mecanum wheels, either arranged in a standard or non-standard shape, this criteria has been evaluated using either the velocity Jacobian \cite{He2019,Borisov2015}, linearised dynamics \cite{Watson2020}, or by heuristics, using a line-crossing approach \cite{Li2019}. The literature in general lacks experimental validation for non-standard configurations; the experiments in \cite{He2019} evaluated the X- (standard) and O-configuration (inverted standard) only.

\section{Hardware Design}\label{s:hardware}

\begin{figure}
    \includegraphics[width=0.5\textwidth]{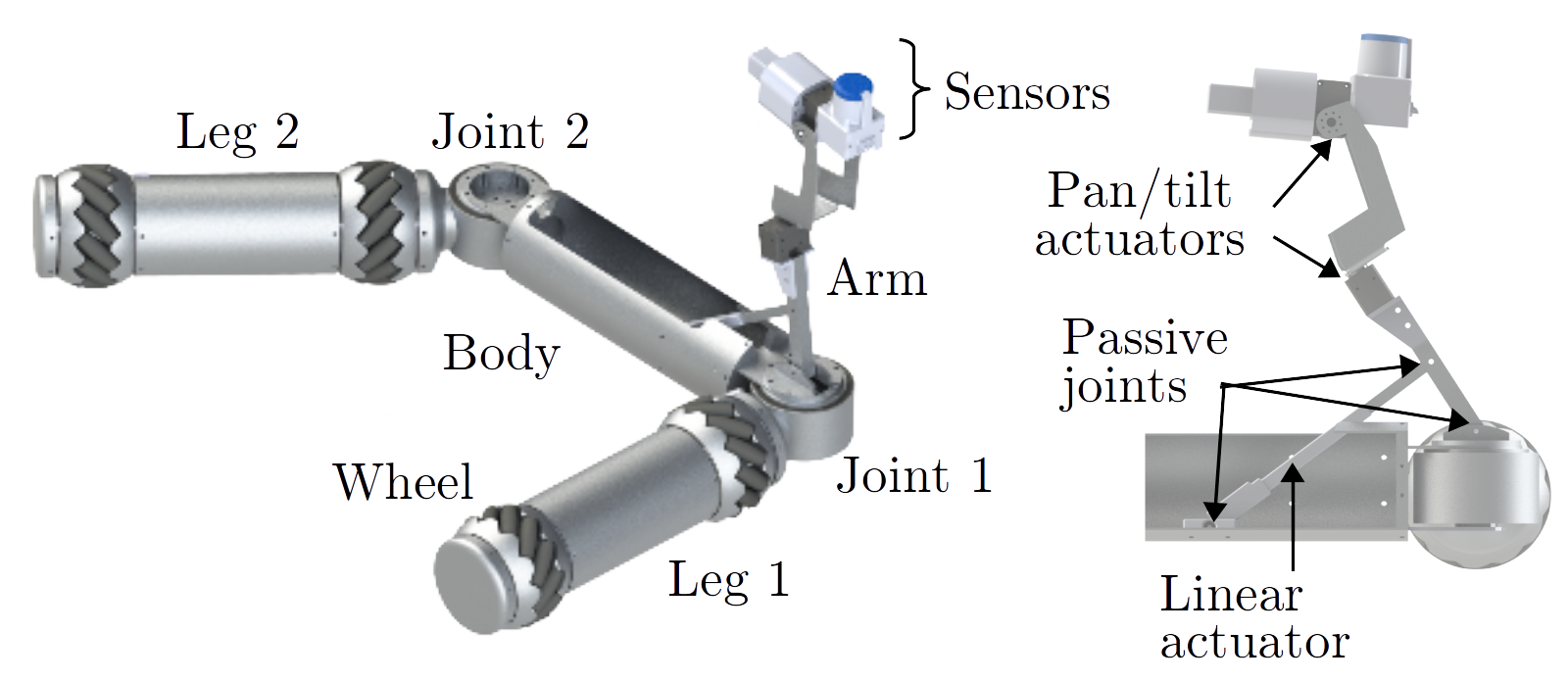}
    \caption{(\textit{left}) The main elements of the MIRRAX: body, legs, wheel and joint assemblies (\textit{right}) Arm assembly.}
    \label{fig:mirrax_hardware}
\end{figure}

MIRRAX consists of a tubular body and two tubular legs, with a pair of wheels connected to each leg and joints connecting the legs to either end of the body section. Fig.~\ref{fig:mirrax_hardware}~shows the main components of MIRRAX while Table~\ref{tb:spec} summarises the technical specification of the robot. The joint assemblies allow the legs to rotate about their connection to the main body through 190\degree in the~$x-y$ plane. The legs of MIRRAX house the robot's electronics and the body section carries the sensor payload on a pan-tilt unit attached to a linear-actuated lever arm which can be raised or lowered. (see Fig.~\ref{fig:mirrax_hardware}\textit{right}). 

The design of the arm enables a larger payload mass to be attached to its end compared to an articulated arm.
Having the capability to raise and lower the arm serves two purposes. First, the overall height of the robot can be adjusted for the 150~mm port ingress or overhanging obstacles. 
Secondly, the direction and height of the sensor can be adjusted for improving the quality of data collected. 

For robustness, MIRRAX's body, legs, joints and wheel hubs are constructed from custom made aluminium parts. The hollow Mecannum wheels are also custom made, using 3D printed parts. 

\begin{table}
\centering
\caption{General specifications of the MIRRAX robot}\label{tb:spec}
\begin{tabular}{lll}\hline
Description             & \multicolumn{2}{l}{Value}                                                                    \\ \hline
External dimensions     & \multicolumn{2}{l}{(L x W x H)} \\
(U-configuration)       & \multicolumn{2}{l}{0.58~m x 0.51~m x 0.13~m} \\
(Straight-line configuration) & \multicolumn{2}{l}{0.140~m x 0.13~m x 0.13~m} \\
                        & Body + Arm                                     & 3.1~kg                                 \\
                        & Leg 1                                          & 0.7~kg                                 \\
Weight & Leg 2                                          & 0.8~kg                                 \\
                        & Wheel (each)                                   & 1.15~kg                                \\
                        & Joint (each)                                   & 0.3~kg                                 \\
                        & \textit{Total}                                 & 9.8~kg                                 \\
Payload size            & \multicolumn{2}{l}{Up to 80~mm x 240~mm}                                           \\
Payload weight          & \multicolumn{2}{l}{Max 2kg}                                                                  \\
Communication interface & \multicolumn{2}{l}{Wireless or tethered}                                                     \\
Battery/run-time        & \multicolumn{2}{l}{2 hours}                                                                  \\
Maximum velocity        & \multicolumn{2}{l}{0.78~m/s}            \\\hline
\end{tabular}
\end{table}

\subsection {Locomotion and Articulation}
\begin{figure}
    \includegraphics[width=0.5\textwidth]{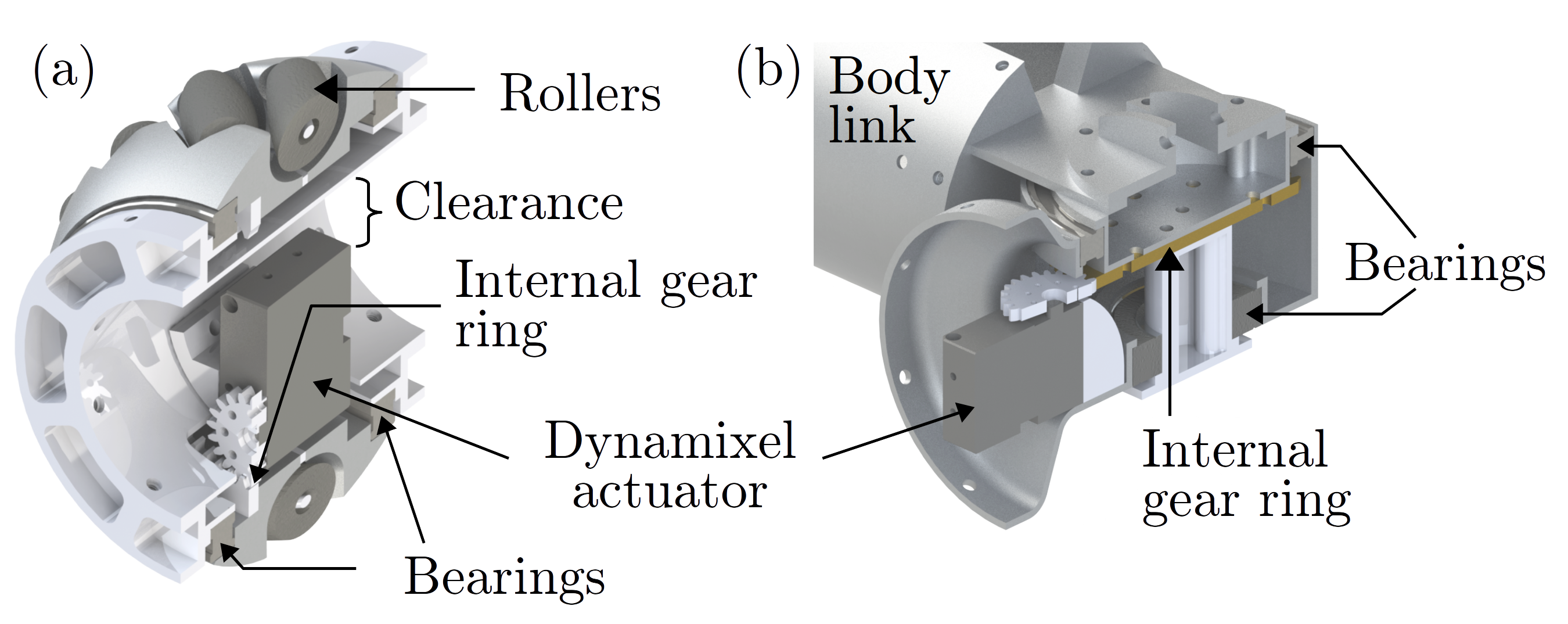}
    \caption{Sectional view of (a) wheel, and (b) joint assembly.}
    \label{fig:cad_sectional}
\end{figure}

MIRRAX's Mecannum wheels run on bespoke hollow axle hubs, allowing space for the servo~motors to sit inside the wheels and for the wiring loom to run through the centre of the robot (see Fig.~\ref{fig:cad_sectional}a).
The wheels are connected to the hub via thin profile sealed roller bearings.
MIRRAX's wheels are driven by  velocity controlled servo~motors with toothed gears attached to their horns.
The toothed gear drives an internal ring gear that is fixed to the inner radius of the wheel with a 3:1 reduction ratio.
The maximum linear velocity of MIRRAX is 0.78\,m/s\textsuperscript{2}; this limit is a result of the maximum velocity of the Dynamixel~XM430\=/350T servo~motors used, and the requirement for hollow wheels leading to the use of a ring gear.
Despite the low maximum velocity, these particular servo~motors were selected for their size, power consumption, reliability and ease of integration.

The joints of MIRRAX are actuated in a similar way, using matching servo~motors and an internal ring gear (see Fig.~\ref{fig:cad_sectional}b).
The key difference is that the servo~motors in the joints are operated in position control mode rather than velocity control to enable the joint angles to be regulated.
The gearing ratio between the Servo~motor and the joint is also 3:1.

\subsection{Electronics Architecture}

\begin{figure} [hb]
    \includegraphics[width=8.5cm]{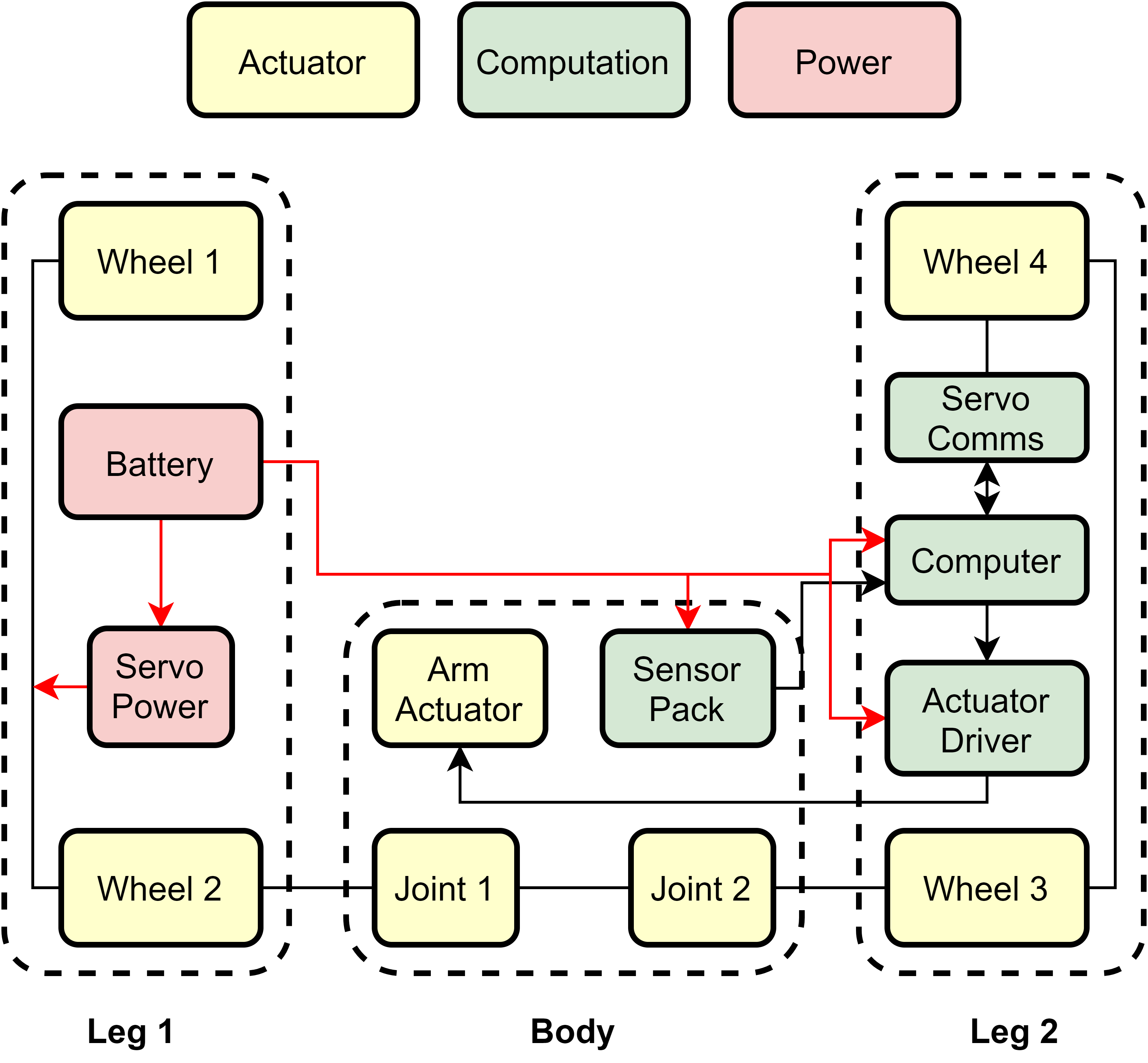}
    \caption{Schematic showing the system architecture of MIRRAX's electronic components.}
    \label{fig:mirrax_arch}
\end{figure}

Fig.~\ref{fig:mirrax_arch} shows a schematic of the layout of, and interconnections between, the electronic components inside MIRRAX. A description of the electronic architecture follows.

MIRRAX is powered by a 4\,Ah~11.1\,V lithium polymer battery housed in Leg 1. At full charge, the battery capacity enables the robot to be operated for approximately 2~hours. Alternatively, the robot can be operated indefinitely using a tether for both power and communication. Alongside the battery, Leg 1 houses the power board for the servo~motors, on/off switches and the charging port.

All computational components are housed in Leg 2.  The main computation unit is a single board PC running Linux and ROS (Robot Operating System). All of the software for MIRRAX is written in C++ and is deployed within the ROS architecture.
All software runs on the single board PC inside MIRRAX, any external connected computers are used for sending simple messages such as position estimates, waypoints or velocity commands if MIRRAX is under manual control.
In addition to the single board PC, Leg 2 houses the servo communication board, the actuator driver for the arm and the WiFi antennas. MIRRAX can connect to external control computers using either WiFi or Ethernet via a tether.

The body element of MIRRAX is used to store the sensor payload, this will include both sensors for navigation and any sensors that will be used for surveying or monitoring tasks.
Sensors in the body element are attached to an articulated arm that rises vertically from the body section.
The default sensor package is a rotating LiDAR for navigation and 3D reconstruction paired with a collimated cadmium zinc telluride (CZT) gamma\=/ray and x-ray detector.

\section{Robot Modelling and Control}\label{s:modelling}

\begin{figure}[ht]
	\begin{centering}
		\includegraphics[width=9cm]{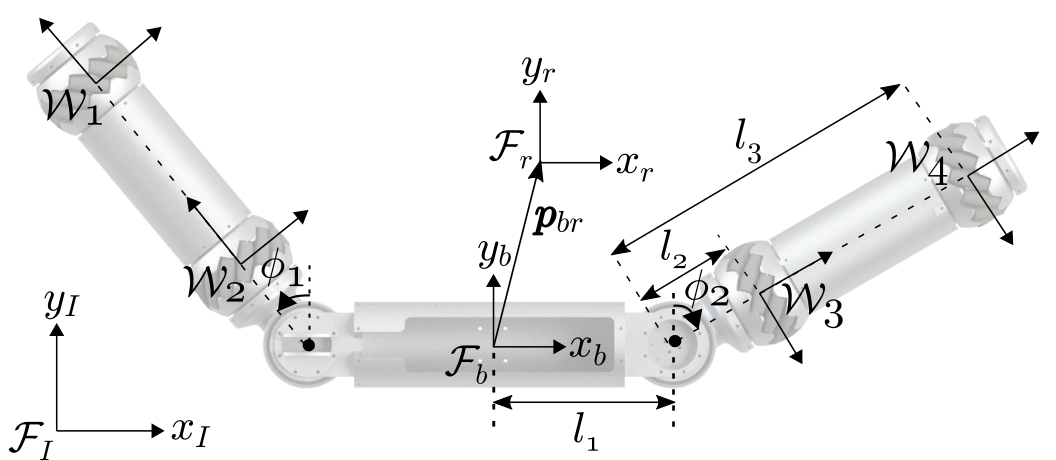}
		\par\end{centering}
	\caption{MIRRAX schematic showing all frames and joints.}
	\label{fig:mirrax_frames}
\end{figure}

The derivation of the kinematic and dynamic model of a reconfigurable mobile robot with Mecanum wheels here extends upon the well established literature on their fixed configuration counterpart \cite{He2019,Borisov2015,Watson2020,Li2019,Agullo1987}. The annotations for the MIRRAX robot are shown in Fig.~\ref{fig:mirrax_frames}. The inertial and base frame are labelled as $\mathcal{F}_{I}$ and $\mathcal{F}_{b}$ respectively. The frame, $\mathcal{F}_{r}$, corresponds to the robot's geometrical centre between the wheels. The wheel frames on each leg link are labelled $\mathcal{W}_{1}$ to $\mathcal{W}_{4}$.
The robot has two leg links attached to its base link via joints $\phi_1$ and $\phi_2$.

The generalised coordinates of the robot is defined as
\begin{equation*}
\bm{q} = 
\left[
p_x , p_y , \theta , \phi_{1} , \phi_{2} , \sigma_1 , ... , \sigma_{n_w} , \psi_1 , ... , \psi_{n_w}
\right]^T
\end{equation*}
where the first three terms are the $x$ and $y$ coordinates of the base, and the rotation angle $\theta$ between the axes $x_b$ and $x_I$, $\phi$ is the leg's joint angle, $\sigma_i$ and $\psi_i$ are the wheel and roller angular position respectively for wheel $i = \{1, ..., n_w\}$, where $n_w$ is the number of wheels of the robot, in this case $n_w=4$. A more compact form for the reduced generalised coordinates for the robot is expressed as
\begin{equation*}
    \bm{x} = \left[p_x, p_y,\theta, \phi_1, \phi_2\right]^T
\end{equation*}

The robot is actuated by changing either the velocity or torque for each wheel and leg joints. Therefore the control input vector for velocity and torque control are defined respectively as
\begin{align*}
    \bm{u}_v &= \left[\dot{\sigma_1}, \dot{\sigma_2}, \dot{\sigma_3}, \dot{\sigma_4}, \dot{\phi_1}, \dot{\phi_2} \right]^T, \\
    \bm{u}_{\tau} &= \left[\tau_{\sigma_1}, \tau_{\sigma_2}, \tau_{\sigma_3}, \tau_{\sigma_4}, \tau_{\phi_1}, \tau_{\phi_2} \right].
\end{align*}

\subsection{Velocity Mapping}
The velocity mapping between the reduced generalised velocities, $\dot{\bm{x}}$, and the wheel velocity, $\dot{\sigma}$, is first derived followed by the full mapping between $\dot{\bm{x}}$ and $\bm{u}_v$.

The no-slip and rolling constraints of the roller's point contact, $c$, as shown in Fig.~\ref{fig:wheel_frame}, for an arbitrary wheel, is expressed as follows:

\begin{figure}[ht]
	\begin{centering}
		\includegraphics[width=8cm]{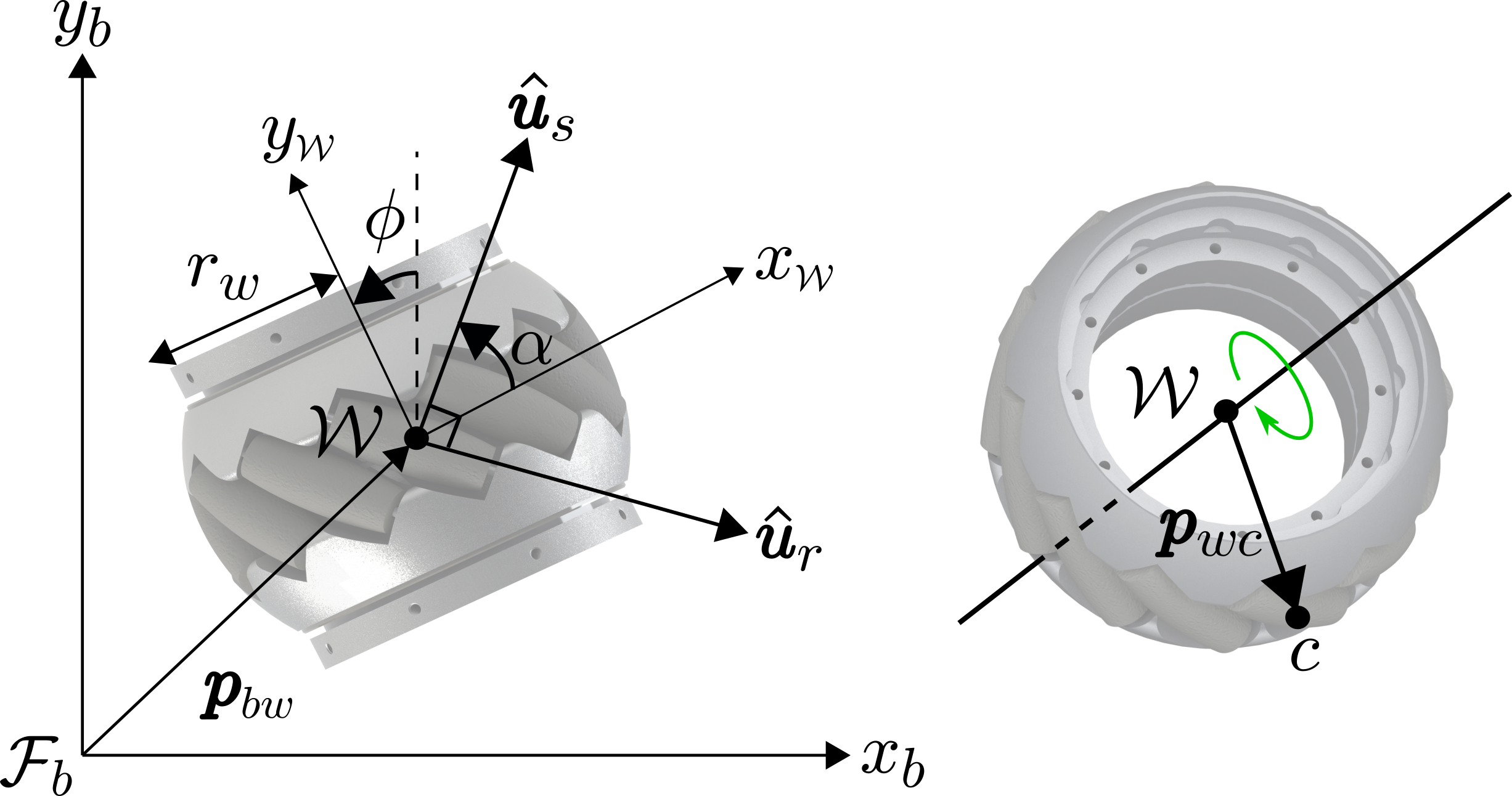}
		\par\end{centering}
	\caption{Wheel frame coordinates and parameters. (Left) Top view (Right) Non-isometric 3D view.}
	\label{fig:wheel_frame}
\end{figure}

\begin{align}\label{eq:kin_constraint_1}
    {}_{w}\bm{v}_{c} \cdot \bm{\hat{u}}_{s} &= 0 \\\label{eq:kin_constraint_2}
    {}_{w}\bm{v}_{c} \cdot \bm{\hat{u}}_{r} &= \dot{\psi} r_{r} 
\end{align}
where ${}_{w}\bm{v}_{c} \in \mathbb{R}^3$ is the linear velocity at $c$, $r_r$ is the roller radius, $\bm{\hat{u}}_{s} \in \mathbb{R}^3$ and $\bm{\hat{u}}_{r} \in \mathbb{R}^3$ are the unit vector, expressed in the wheel frame, for the no-slip and rolling direction respectively. For generality, the sub-indices have been excluded in Fig.~\ref{fig:wheel_frame}.

The linear velocity at the wheel frame, $\mathcal{W}$, due to both the base frame linear and angular velocity, and the leg joint angular velocity, $\dot{\phi}$, expressed in the robot's base frame and wheel frame respectively, is defined as
\begin{equation}\label{eq:v_base_wheel}
{}_{b}\bm{v}_w = 
\bm{R}^T_z(\theta)
\begin{bmatrix}
\dot{p_x} \\ \dot{p_y} \\ 0
\end{bmatrix}
+
\begin{bmatrix}
0 \\ 0 \\ \dot{\theta}
\end{bmatrix} \times \bm{p}_{bw}(\phi)
+
\begin{bmatrix}
0 \\ 0 \\ \dot{\phi}
\end{bmatrix} \times
\bm{R}_z(\phi) \bm{p}_{\phi w}
\end{equation}
\begin{equation}
    {}_{w}\bm{v}_w = \bm{R}^T_z(\phi) \cdot {}_{b}\bm{v}_w
\end{equation}
where $\bm{R}_z(.) \in SO(3)$ is the rotation matrix about the z-axis, $\bm{p}_{bw}(\phi) \in \mathbb{R}^3$ is the position vector from the base to wheel frame, and $\bm{p}_{\phi w} \in \mathbb{R}^3$ is the position vector from the leg joint to the wheel frame.

The linear velocity at $c$ due to the wheel angular rotation, $\dot{\sigma}$, and the linear velocity from the base frame, ${}_{w}\bm{v}_w$, expressed in the wheel frame, is given by
\begin{equation}\label{eq:wheel_vel}
{}_{w}\bm{v}_c = {}_{w}\bm{v}_w + \dot{\sigma} \bm{\hat{e}}_y \times \bm{p}_{wc}
\end{equation}
where $\bm{\hat{e}}_y$ is the unit vector in the y-axis.

Substituting \eqref{eq:wheel_vel} into both \eqref{eq:kin_constraint_1} and \eqref{eq:kin_constraint_2}, and rearranging yields
\begin{equation}\label{eq:ik_wheel_1}
    \begin{bmatrix}
    \dot{\sigma} \\ \dot{\psi}
    \end{bmatrix}
    =
    \begin{bmatrix}
    \bm{d}_w \\ \bm{d}_r
    \end{bmatrix}
    \dot{\bm{x}}
\end{equation}
where $\bm{d}_w$, $\bm{d}_r \in \mathbb{R}^{1\times5}$.

The mapping from $\dot{\bm{x}}$ to the wheel and roller angular velocities explicitly requires the wheel attached to the the leg joint to actively rotate when the leg joint moves to prevent dragging.

Stacking \eqref{eq:ik_wheel_1} for each wheel and rearranging yields

\begin{equation}\label{eq:jacobian}
    \begin{bmatrix}
    \dot{\sigma_{1}}, ..., \dot{\sigma_{4}}, \dot{\psi}_{1}, ..., \dot{\psi}_{4} \\ 
    \end{bmatrix}^T 
    = 
    \begin{bmatrix}
    \bm{D}_{w} \\
    \bm{D}_{r}
    \end{bmatrix}
    \dot{\bm{x}}
\end{equation}
where $\bm{D}_{w}, \bm{D}_{r} \in \mathbb{R}^{4\times5}$ (see Appendix A for the full equation). 

Since the rollers are passive and not controllable, only the top half of \eqref{eq:jacobian} is required for the inverse mapping between the velocity control input and the reduced generalised velocities, which is given by
\begin{equation}\label{eq:idk}
    \bm{u}_v = 
    \begin{bmatrix}
    \bm{D}_{w} \\
    \bm{0}_{2\times3} \quad \bm{I}_{2\times2}
    \end{bmatrix}
    \dot{\bm{x}}
    \triangleq \bm{A} \dot{\bm{x}}
\end{equation}
where $\bm{A} \in \mathbb{R}^{6\times5}$.

The inverse mapping in \eqref{eq:idk} is expressed with respect to the base frame, $\mathcal{F}_{b}$. 
It may be desirable to control the robot about its geometrical centre, $\mathcal{F}_{r}$. 
In which case, $\bm{p}_{bw}$ in \eqref{eq:v_base_wheel} is replaced with $\bm{p}_{rw} \triangleq \bm{p}_{rw}(\phi)$, which is the position vector from $\mathcal{F}_{r}$ to $\mathcal{W}$, defined as 
\begin{align}
    \bm{p}_{br} &= \frac{1}{4}\sum_{i=1}^{4} \bm{p}_{bw,i} \\
    \bm{p}_{rw} &= \bm{p}_{bw} -\bm{p}_{br}.
\end{align}

The direct mapping between the velocity control input and the reduced generalised velocities is
\begin{equation}\label{eq:fdk}
    \dot{\bm{x}} = \bm{A}^+ \bm{u}_v
\end{equation}
where $(.)^+$ is the Moore-Penrose Pseudoinverse. 
Since $\bm{A}$ has full column-rank, then $\bm{A}^+$ has full row-rank, which implies that the direct mapping \eqref{eq:fdk} is under-determined thus have infinite solutions given by the set
\begin{equation}\label{eq:forms_min}
    X = \left\{ \dot{\bm{x}} \in \mathbb{R}^n : \dot{\bm{x}} = \bm{A}^+\bm{u}_v + \left(\bm{I} - \bm{A}^+\bm{A} \right) \bm{z} \right\}.
\end{equation}
The set \eqref{eq:forms_min} includes solutions for the reduced generalised velocity $\dot{\bm{x}}$ that are in the nullspace of $\bm{A}^+$. Since only solutions in the range space of $\bm{A}$ are kinematically feasible,  \eqref{eq:fdk} should only be used after ensuring that $\bm{u}_v$ is in the range space of $\bm{A}$.

\subsection{Dynamics}
Due to the non-holonomic constraints, the equation of motion has been derived using the Euler-Lagrange of the second kind, namely with Lagrange multipliers for MIRRAX. The equation of motion takes the form
\begin{equation}\label{eq:eom_general}
    \bm{M}(\bm{q})\Ddot{\bm{q}} + 
    \bm{C}(\bm{q},\Dot{\bm{q}})\Dot{\bm{q}}
    = 
    \bm{B}\bm{u}_\tau + 
    \bm{\Lambda}(\bm{q})^T\bm{\lambda} + 
    \bm{Q}(\bm{q})\bm{\Dot{q}}.
\end{equation}

The mass-inertia matrix, $\bm{M}(\bm{q}) \in \mathbb{R}^{13\times13}$, is derived using the kinetic energy of the system. 
The Centripedal and Coriolis term, $\bm{C}(\bm{q},\Dot{\bm{q}})  \in \mathbb{R}^{13\times13}$, is derived using Christoffel symbols of $\bm{M}(\bm{q})$. The friction terms are grouped together in $\bm{Q}(\bm{q}) \in \mathbb{R}^{13\times 13}$, while $\bm{B} \in \mathbb{R}^{13\times6}$ is a mapping matrix for the actuated joints. 
The Pffafian constraint matrix, $\bm{\Lambda}(\bm{q}) \in \mathbb{R}^{8\times13}$, is obtained by rearranging \eqref{eq:jacobian} to take the form of $\bm{\Lambda}(\bm{q})\bm{\Dot{q}} = \bm{0}$. Finally, $\bm{\lambda} \in \mathbb{R}^8$ is the Lagrange multipliers from the rolling and no-slip constraints for each wheel. 

The Lagrange multipliers are eliminated using the null-space of $\bm{\Lambda}(\bm{q})$, defined as $\bm{N} \in \mathbb{R}^{13\times 5}$ where $\bm{\Lambda}(\bm{q})\bm{N}=\bm{0}$. Multiplying \eqref{eq:eom_general} with $\bm{N}$ as appropriate will result in the reduced generalised coordinate system, $\bm{x}$, where
\begin{align*}
    \bm{M}(\bm{x}) &= \bm{N}^T \bm{M}(\bm{q}) \bm{N} \\
    \bm{C}(\bm{x},\Dot{\bm{x}}) & = \bm{N}^T \bm{C}(\bm{q},\Dot{\bm{q}}) \bm{N} + 
    \bm{N}^T \bm{M}(\bm{q}) \bm{\dot{N}} \\
    \bm{B}_{x} &= \bm{N}^T \bm{B} \\
    \bm{Q}_{x}(\bm{x}) &= \bm{N}^T \bm{Q}(\bm{q}) \bm{N} 
\end{align*}
resulting in
\begin{equation}\label{eq:eom_reduced}
    \bm{M}(\bm{x})\Ddot{\bm{p}} + 
    \bm{C}(\bm{x},\Dot{\bm{x}})\Dot{\bm{x}} + 
    = 
    \bm{B}_{x}\bm{u}_\tau + 
    \bm{Q}_{x}\dot{\bm{x}}
\end{equation}
which can be solved for the forward dynamics,
\begin{equation}\label{eq:for_dyn}
    \Ddot{\bm{p}} =
    (\bm{M}(\bm{x}))^{-1}
    \left[
    \bm{B}_{x}\bm{u}_\tau + 
    \bm{Q}_{x}\dot{\bm{x}} -
    \bm{C}(\bm{x},\Dot{\bm{x}})\Dot{\bm{x}}
    \right].
\end{equation}

\subsection{Control System}\label{ss:control}
To enable trajectory tracking, a feedback controller was employed for the robot's base velocity. The controller velocity output, 
$\dot{\bm{x}}_{c} \in \mathbb{R}^5$, for a desired position, $\bm{x}_{d}$ and velocity, $\dot{\bm{x}}_{d}$, is described by the control law
\begin{equation}\label{eq:base_pd}
    \dot{\bm{x}}_{c} = \dot{\bm{x}}_{d} + \bm{K}_{p}(\bm{x}_{d}-\bm{x})
\end{equation}
where $\bm{K}_{p} \in \mathbb{R}^{5\times5}$ is diagonal gain matrix.

The resulting $\bm{u}$ from \eqref{eq:idk} using $\dot{\bm{x}}_{c}$ in \eqref{eq:base_pd} may not be within the actuator limits. It is not possible to directly impose fixed velocity limits on $\dot{\bm{x}}_{c}$ since the velocity limits are configuration dependant. Furthermore, it is desirable to have the robot travel at a feasible velocity, $\dot{\bm{x}}_{f}$, parallel to $\dot{\bm{x}}_{c}$, i.e. $\dot{\bm{x}}_{f} = \beta \dot{\bm{x}}_{c}$, where
$\beta \in [0,1)$.
The above requirement is achieved by solving the following quadratic programming problem at each time step,
\begin{equation}
\begin{aligned}
    \dot{\bm{x}}_{f} \in \arg & \min_{\dot{\bm{x}}}      \quad  \|\dot{\bm{x}} - \dot{\bm{x}}_{c}\|^2 \\
    \textrm{s.t.} & \quad -\bm{u}_{lim}  \leq \bm{A} \dot{\bm{x}} \leq \bm{u}_{lim}\\
     & \quad \quad \dot{\bm{x}} \times \dot{\bm{x}}_{c} = \bm{0}   \\
\end{aligned}
\end{equation}
where $\bm{u}_{lim} \in \mathbb{R}^6$ is the actuator velocity limit.

\section{Controllability}\label{s:controllability}
The controllability of a re-configurable robot with Mecanum wheels is analysed here using the velocity Jacobian and the linearised dynamics.

\subsection{Velocity Jacobian}\label{ss:contr:jac}
In this paper, the approaches taken in \cite{He2019} and \cite{Borisov2015} are adopted to analyse the controllability of MIRRAX. In \cite{He2019}, a quantitative measure of how good the controllability is provided using the Global Stiffness Index (GSI), $n_{c}$, \cite{Liu2000}

\begin{equation}\label{eq:gsi}
    n_{c} = \frac{1}{\left\Vert \bm{C} \right\Vert \cdot \left\Vert \bm{C}^{-1} \right\Vert}
\end{equation}
where
\begin{equation*}
    \bm{C} = (\bm{W})^T \cdot \bm{W}, \quad
\left\Vert \bm{C} \right\Vert = \sqrt{\Tr \left(\frac{1}{3} \bm{C} \bm{C}^{T}\right)},
\end{equation*}
and $\bm{W}$ is the first three columns of the matrix $\bm{D}_w$.

A larger GSI would result in better disturbance rejection and better trajectory tracking, whereas having a zero GSI means the system is uncontrollable. Similarly, the authors in \cite{Borisov2015} use the condition of having a non-zero determinant as a criteria for controllability, provided that at least three wheels are in contact, defined as
\begin{equation}\label{eq:borisov}
d_{c} = \text{det}
\begin{bmatrix}
\alpha_{11} & \alpha_{21} & \alpha_{31} \\
\alpha_{12} & \alpha_{22} & \alpha_{32} \\
(\bm{J}\bm{p}_{1} \cdot \bm{\alpha}_{1}) &
(\bm{J}\bm{p}_{2} \cdot \bm{\alpha}_{2}) &
(\bm{J}\bm{p}_{3} \cdot \bm{\alpha}_{3}) \\
\end{bmatrix}
\neq 0
\end{equation}
where $\bm{p}_{i}$ is the roller position relative to the robot frame, $\bm{\alpha}_{i}$ is the roller axis direction, and
\begin{equation*}
    \bm{J} = \begin{bmatrix}
    0 & -1 \\ 1 & 0
    \end{bmatrix}
\end{equation*}

\begin{figure}[htbp]
\includegraphics[width=9cm]{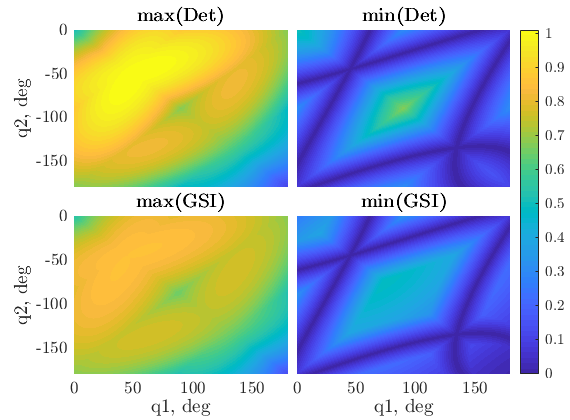}
\caption{Controllability of MIRRAX for different configurations evaluated using the determinant approach (top), and the GSI approach (bottom).}
\label{fig:det_gsi}
\end{figure}

Similar to these approaches, the controllability of MIRRAX was evaluated at fixed leg configurations i.e. the influence of the leg joints on the inverse mapping in \eqref{eq:idk} were removed by eliminating the last two rows and columns of $\bm{A}$.
Fig.~\ref{fig:det_gsi} shows the GSI and the determinant from \eqref{eq:gsi} and \eqref{eq:borisov} for the full range of joint configurations for MIRRAX. The plots Fig.~\ref{fig:det_gsi}a-b shows the maximum while Fig.~\ref{fig:det_gsi}c-d shows the minimum GSI and determinant respectively. The maximum and minimum determinant for a particular configuration was determined among the possible wheel combinations, i.e.
\begin{align}\label{eq:dmin}
    d_{min} &= \min(d_{c1},d_{c2},d_{c3},d_{c4}) \\
    d_{max} &= \max(d_{c1},d_{c2},d_{c3},d_{c4})
\end{align}
where $c1$ to $c4$ correspond to wheels in contact in the following combination (1,2,3), (1,2,4), (1,3,4) and (2,3,4) respectively.
The GSI approach differs slightly in evaluating \eqref{eq:dmin} with the addition of the wheel combination $c5$, where all the wheels are in contact.

It can be observed from all four plots in Fig.~\ref{fig:det_gsi} that the robot's controllability is symmetrical about the top-left to bottom-right diagonal axis. This indicates symmetrical relationship for configurations about the robot's lateral axis, e.g. for an arbitrary leg joint angle $\phi_{a}$ and $\phi_{b}$, the controllability is the same, $n_{c}(\phi_{a},\phi_{b}) = n_{c}(\phi_{b},\phi_{a})$.

Comparing between Fig.~\ref{fig:det_gsi}a-b and Fig.~\ref{fig:det_gsi}c-d, it can be seen that the controllability is dependant on the wheels having ground contact. Under the ideal condition where all the wheels are in contact, the robot is controllable for all joint configurations since both the determinant and GSI are non-zero (Fig.~\ref{fig:det_gsi}a-b). Both these metrics approach zero as $|\phi_{1,2}|\xrightarrow[]{}180\degree$, the inverted-U shape, implying an increase in susceptibility to disturbance and being less controllable i.e. taking longer to reach a desired state.
However, the possible loss in wheel ground contact for extended periods of time can render the robot uncontrollable as seen by the dark blue regions in Fig.~\ref{fig:det_gsi}c-d where the determinant and GSI are zero.

The GSI using the robot's full inverse mapping matrix, $\bm{A}(\bm{x})$, result in $\text{GSI}\approx 0$ at $\phi_{1,2} = 0$ (standard shape configuration), is markedly different from its fixed-configuration counterpart where $\text{GSI} = 0.216$. This result suggest that the GSI approach is not directly adaptable to include reconfigurability in the controllability analysis used above.

\subsection{Linearised Dynamics}\label{ss:contr:kcm}
The global controllability for a non-linear system, as is the case in this study, does not exist as compared to a linear system. A weaker form of this proof is to instead show that it is small-time locally controllable (STLC), as shown in \cite{Watson2020} for a collinear mecanum drive. A similar approach is used here, namely by evaluating the Kalman Controllability Matrix (KCM) using the linearised system dynamics at selected state. The KCM should have full rank for the system at the linearised state to be STLC.

For the state vector, $\bm{z} = [\bm{x}, \bm{\Dot{x}}]^T$, the resulting linearisation takes the form 
\begin{equation}\label{eq:kcm_lin}
    \dot{\bm{z}} = \bm{A}_L\bm{z} + \bm{B}_L\bm{u}_\tau
\end{equation}
where
\begin{equation*}
    \bm{A}_L = 
    \begin{bmatrix}
    \bm{0}_{5\times5} &\bm{I}_{5\times5} \\
    \bm{0}_{5\times5} &\frac{\partial\bm{D}}{\partial \bm{\Dot{x}}}
    \end{bmatrix}, \quad
    \bm{B}_L = 
    \begin{bmatrix}
    \bm{0}_{5\times6} \\
    \frac{\partial\bm{D}}{\partial \bm{u}}
    \end{bmatrix}
\end{equation*}
and $\bm{D}$ is the forward dynamics from \eqref{eq:for_dyn}.

For MIRRAX, the arms configuration of the robot is most likely to affect the controllability of the robot. Hence the robot's controllability was evaluated at equilibrium states corresponding to different arm configurations, similar to Section~\ref{ss:contr:jac}. Stepping through the various joint angle combinations at $5\degree$ intervals and evaluating the KCM of \eqref{eq:kcm_lin} resulted in full rank for all the configurations evaluated, namely $\text{rank}\left(\text{ctrb}[\bm{A_L},\bm{b}_L]\right) = 10$, suggesting that the robot is at least STLC at these evaluated configurations.

\section{Experimental Evaluation and Discussion}\label{s:experiment}
 This section presents experimental validation of the robot in achieving (a) ingress and egress through 150~mm diameter access ports, and (b) controlled navigation of confined areas. A \href{https://tinyurl.com/2kehf43z}{video} showing the experiments is available.  The data supporting the findings reported in this paper are openly available from the FigShare repository at  https://doi.org/10.48420/21071017.

\subsection{Experimental Set Up}
Manual control of the robot's base velocity follows a simple mapping from a PS4 joystick to ($\dot{x},\dot{y},\dot{\theta}$); similar to ROS 2D twist commands. Additional velocity commands and functionalities were introduced to actuate the legs and arms (integrating the velocity commands for the legs and arm positions) using the buttons available on the PS4 joystick. 

For trajectory tracking, a 5D trajectory of $\bm{x}$ was generated using an adapted version of the ROS mav\_trajectory\_generation\footnote{https://github.com/ethz-asl/mav\_trajectory\_generation} package. Feasibility of the trajectory generated against the wheel velocity limit was checked in an exhaustive manner using \eqref{eq:jacobian}. In the event that the trajectory violates the wheel velocity limits, the trajectory was scaled in time by the ratio $s = \omega/\omega_{max} + 0.05$ and the feasibility check repeated until it becomes feasible.
A VICON system was used for both ground truth and the feedback controller, which runs at 100~Hz.

\subsection{Limited Access Entry}\label{ss:exp:access}
\begin{figure}[htbp]
    \centerline{\includegraphics[width=9cm]{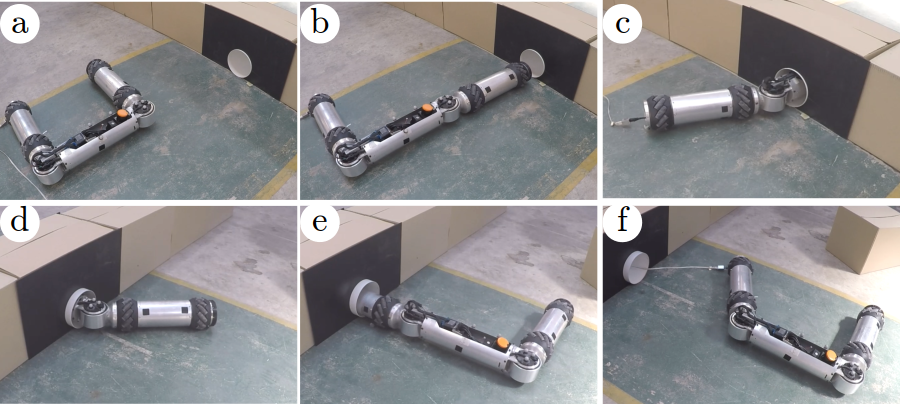}}
    \caption{Ingress and egress from a 150~mm diameter access port.}
    \label{fig:ingress}
\end{figure}

The primary hardware requirement is for the robot to ingress and egress 150~mm diameter access ports. A snapshot of this motion is shown in Fig.~\ref{fig:ingress}. 
The robot was first re-configured to a L-shape configuration (Fig.~\ref{fig:ingress}b), then driven down the access port. 
As the rear leg approaches the port, it was moved to $\phi_{1}=90\degree$ (Fig.~\ref{fig:ingress}c). At the same time, the front leg which has emerged from the port was moved towards its default position as much as possible without colliding with the wall (Fig.~\ref{fig:ingress}d).
As the robot continues moving and exits the port completely, the rear leg takes back its default position since the exit area had sufficient clearance.

A scenario that may occur inside or on exiting the port is for the robot to roll around its x-axis. In its current state, it is possible to orientate the robot to be right-side up ($\theta_{roll}=0\degree$) if $-90\degree \lessapprox \theta_{roll} \lessapprox 90\degree$. The robot becomes unable to orientate itself beyond this range due to the placement of components which results in mass imbalance. As such, the counter-torque from the wheels is unable to rotate the robot's base back up. 

\subsection{Mass Balance}\label{ss:exp:mass}
The inherent design of MIRRAX, having both the front and rear leg connected via a link that does not pass through the centre of the robot, results in an off-balance Centre-of-Mass (CoM). The robot's CoM is not fixed due to the robot's reconfigurable capability, thus depending on both the legs and arm position. 
Taking the U-shape (default) configuration as a starting point, the robot's CoM can be calculated from the mass of the individual links, summarised in Table~\ref{tb:spec}. The CoM lateral offset is found to be located to the right of its geometrical centre, while the longitudinal offset is approximately coincident. 

\begin{figure}[htbp]
    \centerline{\includegraphics[width=8.8cm]{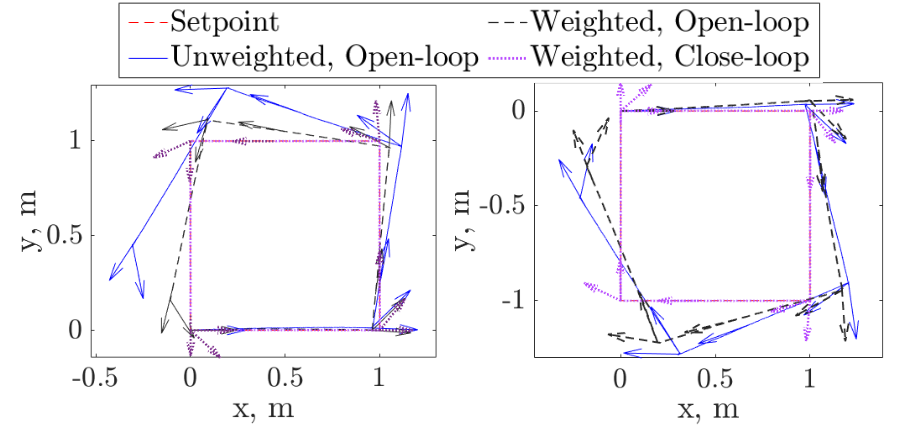}}
    \caption{Trajectory tracking of a square at (0,-0) configuration in open-loop and close-loop with and without counter-mass (a) counter-clockwise and (b) clockwise rotation.}
    \label{fig:mass_0_0}
\end{figure}

\begin{figure}[htbp]
    \centerline{\includegraphics[width=9cm]{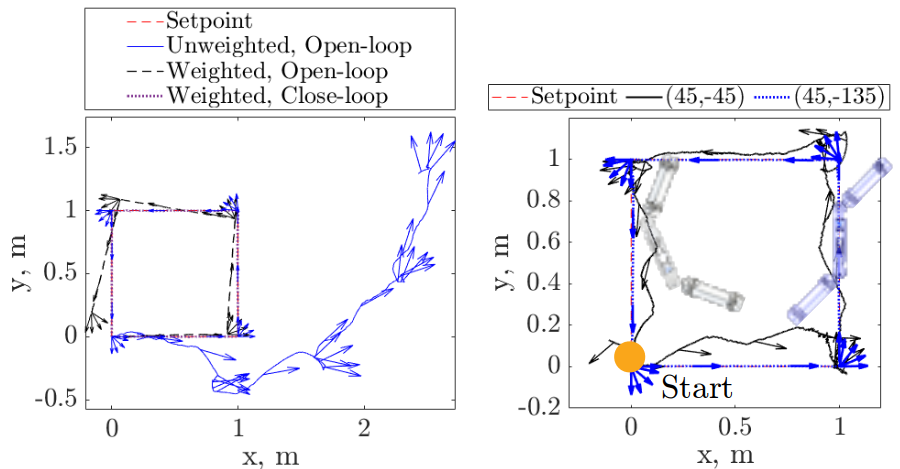}}
    \caption{Trajectory tracking at non-standard a configuration (a) (45,-45) in open-loop and close-loop with and without counter-mass. (b) (45,-45) and (45,-135) close-loop without counter-mass.}
    \label{fig:mas_45_45}
\end{figure}

Offset in the robot's CoM has the undesirable effect of reducing surface traction on one or more wheels, causing the robot to deviate from the desired motion, as shown by the non-symmetrical motion of clockwise and counter-clockwise motion in Fig.~\ref{fig:mass_0_0} for the default configuration ($\phi_{1} = \phi_{2} = 0\degree$) \cite{Borenstein1995}. The problem becomes more evident at other configurations, e.g. at $\phi_{1} = -\phi_{2} = 45\degree$ where the robot becomes incapable of straight line motion since both the left most wheels lost a sufficient amount of traction, as shown in Fig.~\ref{fig:mas_45_45}a (Unweighted + open-loop).

To address the robot's off-centred CoM, counter-balance mass was attached to the end of the leg links. The location for the mass attachment was selected due to the limited space available inside the leg links, and the end of the leg having the largest distance from the robot's geometrical centre i.e. a larger moment arm reduces the mass required. 
The total mass required for the robot's CoM to be coincident with the geometrical centre was calculated to be 5.44~kg, or 2.72~kg per leg. For the maximum arm payload of 2~kg, a counter-balance mass of 4.24~kg per leg would be required.  

The result of having the mass balance is shown in both Fig.~\ref{fig:mass_0_0} and \ref{fig:mas_45_45}(left) with the legend \textit{weighted} where significant improvement is observed on both open- and close- loop tracking of the desired path. 
However, the overall mass of the robot is increased significantly, where the counter-mass accounts for 36~\% of the final mass. 
A possible strategy to circumvent or reduce the use of counter-mass used is to utilise a Z-configuration, as seen by the imposed image of MIRRAX in Fig.~\ref{fig:mas_45_45}(right). This configuration indirectly balances the mass of the robot, resulting in the CoM to be coincident with its geometrical centre, provided the mass of both legs are similar. 
The resulting trajectory tracking performance using the non-weighted Z-configuration is comparable with the weighted approach. 

Although close-loop does indeed improve the performance especially when operated without the counter-mass, there are advantages when the open-loop tracks the desired path such as ease of control in manual mode and lower power consumption as since it drifts less. Furthermore, higher accuracy can be obtained from odometry localization as well. 
It should be noted that the balancing mass used in this experiment was temporary since the mass required depends on the sensor payload on the arm, which in turn depends on the mission/research requirement. Hence the counter-mass was left as an ad-hoc solution here and will be modified in the future accordingly.

\subsection{Robot Balance}\label{ss:exp:balance}
As the robot approaches the straight\=/line configuration ($\phi_{1} = -\phi_{2} = 90\degree$), it becomes dynamically unstable and susceptible to rolling over. Although it was possible to employ controllers capable of both balancing and moving the robot at this configuration \cite{Watson2020}, the decision was taken to avoid this configuration during operation, except for the ingress and egress from the access port. This was due to the risk of rolling over during operation, which can cause the robot to be unable to right itself as discussed in Section~\ref{ss:exp:access}, and from crashing, which can have detrimental effect on the sensors located on the arm. 

\begin{figure}[htbp]
    \centerline{\includegraphics[width=8.8cm]{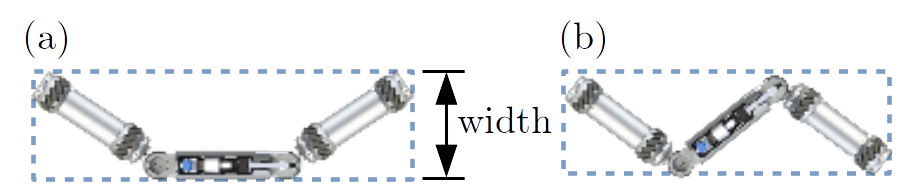}}
    \caption{General configurations for minimum stable footprint (a) trapezoidal (b) Z- shape.}
    \label{fig:roll_config}
\end{figure}

A selected number of configurations were evaluated to characterise the minimum stable footprint using the robot's roll and absolute trajectory error (ATE) as quantitative measures. The configurations explored were generally based on a trapezoidal and Z- shape (see Fig.~\ref{fig:roll_config}): the first when both $|\phi_{1,2}| < 90\degree$ (trapezoidal), the second when $\phi_{1} < 90\degree$ and $\phi_{2} < -90\degree$ (zig-zag).

\begin{figure}[htbp]
    \centerline{\includegraphics[width=9cm]{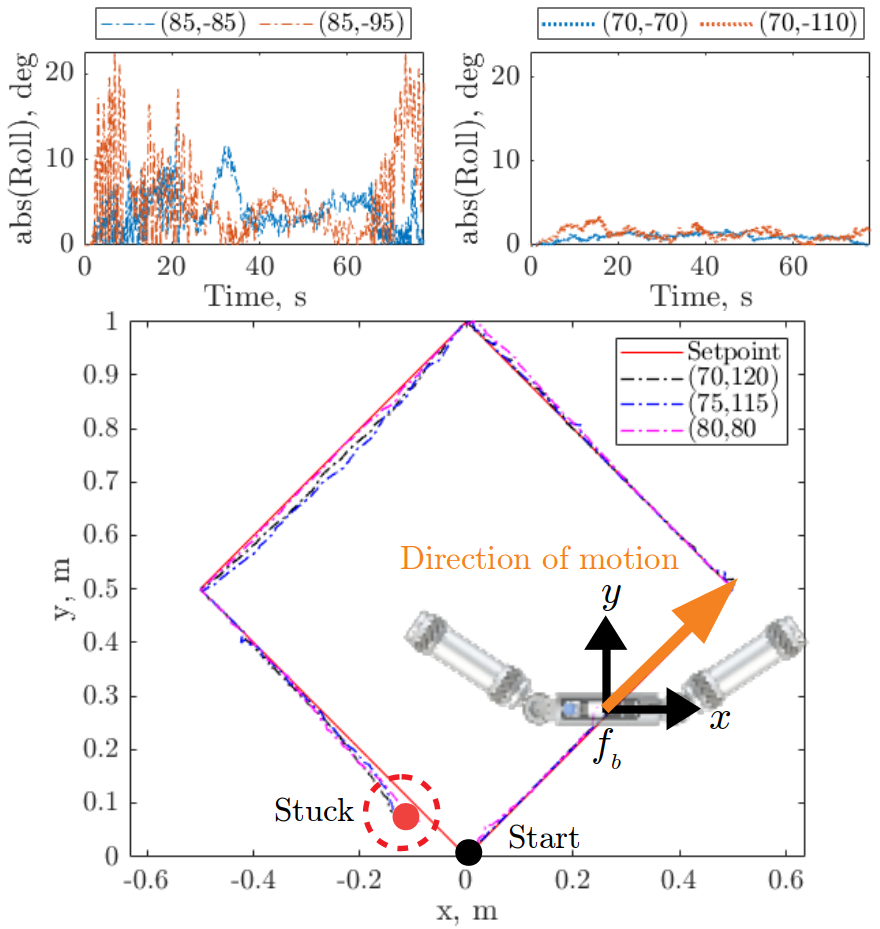}}
    \caption{(\textit{top}) Magnitude of base roll at various configurations for assessing minimum stable footprint. (\textit{bottom}) Trajectory tracking at various configuration experiencing wheel slippage at a similar spot.}
    \label{fig:base_roll}
\end{figure}

Fig.~\ref{fig:base_roll}(\textit{top}) shows the roll angle from executing a motion where the orientation of the robot and inertial frames were fixed and aligned, while the direction of motion was not aligned to the body frame x-axis (see Fig.~\ref{fig:base_roll}(\textit{bottom})).
Table~\ref{tb:roll_summary} summarises the result for the different configurations evaluated.

\begin{table}[htbp]
\centering
\caption{Quantitative assessment of different configurations for minimum stable footprint.}\label{tb:roll_summary}
\begin{tabular}{lllll}\hline
Configuration & Width, m & Max (Roll), $\degree$ & $\sigma$, $\degree$ & ATE, m \\ \hline
(85,-85)      & 0.15     & 13.79  & 3.35     & 0.27   \\
(80,-80)      & 0.19     & 4.82   & 0.84     & 0.17   \\
(75,-75)      & 0.22     & 2.46   & 0.54     & 0.26   \\
(70,-70)      & 0.26     & 1.96   & 0.44     & 0.12   \\
\hline
(85,-95)      & 0.14     & 22.53  & 6.99     & 0.34   \\
(80,-100)     & 0.16     & 15.08  & 2.35     & 0.32   \\
(75,-105)     & 0.17     & 5.68   & 1.20     & 0.22   \\
(70,-110)     & 0.19     & 3.42   & 1.05     & 0.11   \\
(65,-115)     & 0.21     & 3.74   & 0.69     & 0.13   \\
(75,-115)     & 0.19     & 1.93   & 0.69     & 0.16   \\
(70,-120)     & 0.21     & 1.85   & 0.57     & 0.16   \\
\hline
\end{tabular}
\end{table}

It can be observed from Fig.~\ref{fig:base_roll}(\textit{top}) and Table~\ref{tb:roll_summary} that both the magnitude of the undesirable roll and ATE was especially significant near the straight line configuration: (85,-85) and (85,-95). This was due to these two configurations being close to a dynamically unstable configuration, whereby any form of motion would result in rolling both due to gravity and from the actuators' counter-torque.

The roll and ATE reduced significantly for the subsequent configurations, with the largest improvement observed at (80,-80) and (75,-105). For the remaining configurations, only slight improvements in the reduction of roll was observed, although the Z-shape gained a larger improvement in tracking with a smaller ATE, $\leq0.13$~m.
The (75,-75) configuration had a larger ATE error due to significant wheel slippage observed during the motion. It was expected that the ATE would fall between 0.17~m and 0.12~m under no slip conditions.

The Z-shape configurations does not result in better performance compared to the trapezoidal shape at footprint width of less than $0.19$~m. However the performance changed as the width increased, where the Z-shape achieved better tracking and less roll compared to its counterpart.
The minimum $90\degree$ corner path width that MIRRAX can fit through was approximately 0.3~m (horizontal width). Assuming that the path leading up to the corner is of a similar size, the configurations (75,-115) and (70,-120) would yield a better choice. This selection provides larger wall clearance and similar quantitative metric compared to the (70,-70) configuration.

An interesting observation on the minimum footprint experiments was that the robot was sometimes unable to progress along the desired trajectory due to wheel slippage. This scenario was observed to occur at a similar area for the different experiments carried out (see Fig.~\ref{fig:base_roll}(\textit{bottom}) for the stuck region). Increasing the controller gains and leaving it to continue attempting to reach the goal position\footnote{The start and goal position is the same.} failed with it remaining in the same spot. Whilst not shown here, the robot was made to escape this slippage by inducing motions in other directions e.g. sideways then forward instead of a pure diagonal motion. 
This scenario highlights the susceptibility of being stuck due to wheel slippage, similar to other types of wheeled robots. However, the capability for omni-directional motion in this case allows for additional motions to be used to escape from being stuck. 

\subsection{Controllability}\label{ss:exp:contr}
\begin{figure}[htbp]
    \centerline{\includegraphics[width=9cm]{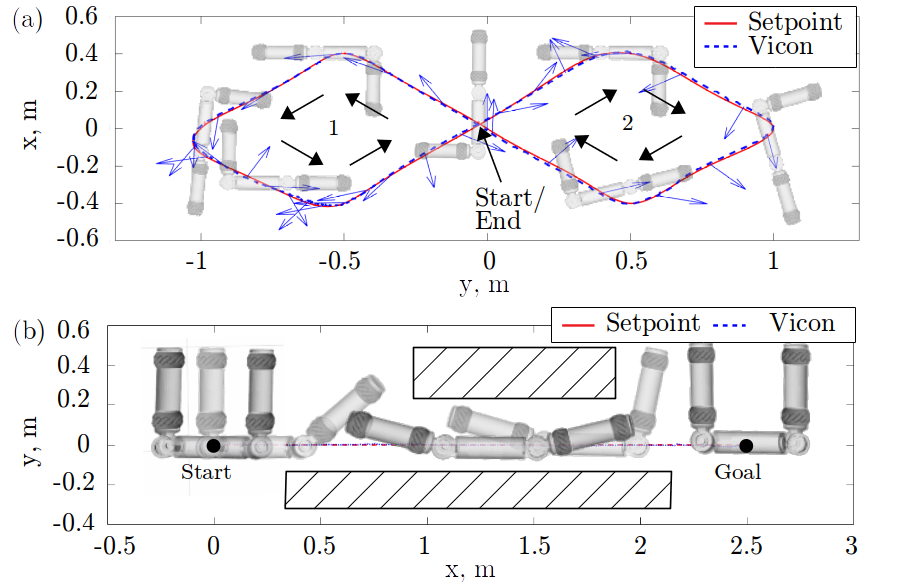}}
    \caption{Trajectory tracking with (a) fixed (b) dynamic leg configurations. }
    \label{fig:contr_demo}
\end{figure}

Using the proofs presented in Section~\ref{s:controllability}, the robot was theoretically controllable over its full range of motion. The practicality of this was evaluated on a number of experiments. 
Fig.~\ref{fig:contr_demo} shows the robot tracking a pre-defined trajectory with either fixed- or dynamic leg configurations. 
Fig.~\ref{fig:contr_demo}a shows a representation of omni-directional motions. Fig.~\ref{fig:contr_demo}b shows the robot navigating through a narrow path.
In general, the robot was able to track the prescribed trajectory closely, where the error magnitude for the xy-position and orientation was less than 0.018~m and 0.6$\degree$ at any time instant for both motions.

\begin{figure}[htbp]
    \centerline{\includegraphics[width=9cm]{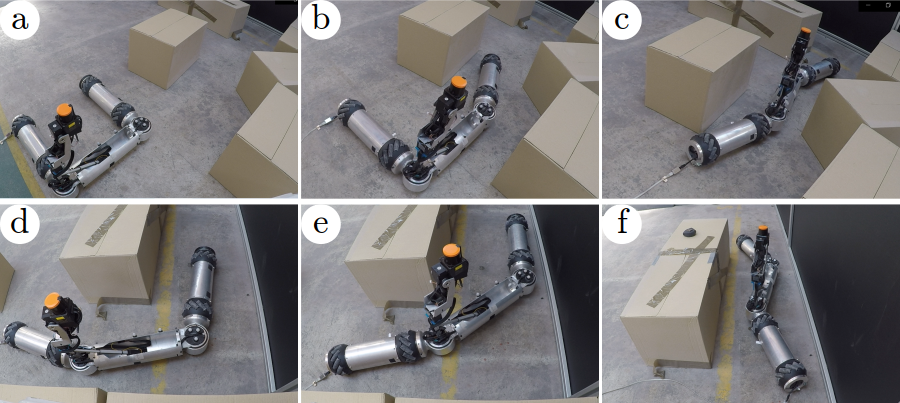}}
    \caption{Manual control demonstration of navigating cluttered and confined environments. }
    \label{fig:manual_demo}
\end{figure}

The robot was further evaluated by attempting to navigate through cluttered and confined environments via manual control mode. Fig.~\ref{fig:manual_demo} shows a snapshot of the robot's motion navigating through a 90$\degree$ corner accessible only through narrow pathway. Control of the robot without the counter-mass was possible and was generally responsive as seen in Fig.~\ref{fig:manual_demo} and the supplementary video. 

There were instances where the wheel lost traction due to the mass balancing issue discussed previously in Section~\ref{ss:exp:mass}. This was observed while navigating the corner which required the robot to take on a configuration close to (45,-45) which was known to cause loss of controllability. However, by perturbing the robot's motion and adjusting the legs it was possible to navigate through the corner. The Z-shape was also used where possible to minimise the robot's roll during motion.

\section{Sellafield Deployment}\label{s:deploy}
A version of the MIRRAX robot was deployed into a real-world, low-level radioactive facility on the Sellafield site, UK in March 2018.  The deployment area was part of the Magnox reprocessing facility which had been sealed for a number of years and was the first time a mobile robot had been deployed into this facility.  The purpose of the deployment was to geometrically characterise the area so that decommissioning plans could be generated.  A video of the deployment is available\footnote{\href{https://tinyurl.com/mvewxx9w}{Deployment video: https://tinyurl.com/mvewxx9w}}.

The robot used in the deployment was largely similar in design with minor differences: the robot's links were constructed from PVC and 3D printed parts and the arm consisted of an actuated articulated joint with a pan unit.
The CoM for the robot used in the deployment was closer to its geometrical centre due to the materials used being lighter. The mass of the wheels was larger than the rest of the links, enabling self-righting from arbitrary roll angles.

The arm design differed from the robot presented earlier in Section~\ref{s:hardware} since the payload used in the deployment was lighter. The sensor payload consisted of two 2D LIDARs attached to a pan unit, enabling 3D mapping while still being able to localise using the forward facing LIDAR. An RGB camera located at the edge of the front leg enabled visual feedback for the operator.  

\begin{figure}[htbp]
    \centerline{\includegraphics[width=9cm]{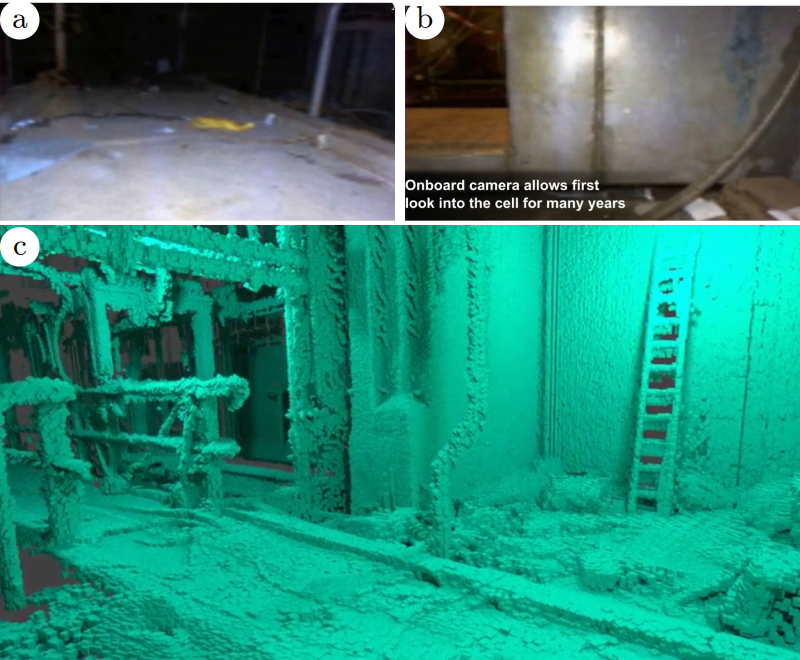}}
    \caption{(a) MIRRAX inside the Magnox facility. (b) RGB camera view (c) rendered 3D point cloud of facility.}
    \label{fig:deploy_mirrax}
\end{figure}

The robot was tethered and operated manually, with the tether being used for both power and communication to the base PC. 
An RGB camera enabled a coloured view of the facility (see Fig.~\ref{fig:deploy_mirrax}a-b).
Inside the facility, the robot utilised its forward-facing LIDAR for 2D SLAM as it was driven forward for inspection and mapping. 
After travelling for short intervals, the robot was stopped and the pan unit was rotated to increase point cloud data collection from the top-facing LIDAR. The point cloud collected was post-processed offline, the final render of the facility is shown in Fig.~\ref{fig:deploy_mirrax}c. 

At the end of the mapping session, the LIDAR units were retrieved and the rest of the robot was disposed as low-level radioactive waste. The overall deployment successfully proved the feasibility of both MIRRAX and in general, mobile robotic platforms, being used for remote inspections in these hazardous environments.

\section{Conclusions \& Future Work}\label{s:conclusion}
This paper has presented MIRRAX, a reconfigurable robot driven by Mecanum wheels. Special focus has been given to its design to enable it to have sufficient payload capability as well as fit through a 150~mm diameter access port. The incorporation of a feedback controller was shown to enable trajectory tracking for arbitrary leg configurations. 
Preliminary experiments confirmed the robot's capability for 150~mm diameter access port entry and omni-directional capability. The validation experiments also highlighted certain limitations, such as the off-balance CoM, wheel slippage, rolling during motion and recovery when MIRRAX operates around the straight-line configuration. The robot was subsequently deployed in the Magnox facility on the Sellafield site, showcasing the relevance of the robot in practical scenarios and providing valuable insight of the facility for the site operators.

The preliminary experiments and deployment highlights a number of challenges remain in both robot development and deployment. 
Although the modelling and controllability analysis of the robot shows the feasibility of such a design in meeting the requirements for restricted access, disturbances to the system which were not modelled, or difficult to model, can affect the actual use of the system. 
Real-world scenarios as seen in the experiment and deployment where the environment has irregular terrain from loose gravel, rocks and obstacles at times result in the robot being stuck, requiring a change in leg configurations or some form of manoeuvre by the tele-operator to escape from the state it is in. 
Indeed, manual control of the 6 DOF simultaneously is very challenging and current teleoperation procedures limit configuration changes to when the robot is static.

The outlook for the robot is to introduce autonomy for operations in unknown environments. Future hardware development will investigate the integration of perception sensors for environment and radiation mapping, both of which are critical for high-level planners and the robot's controller in ensuring the robot's safety and ensuring the desired motion is executed as expected. 
A motion planner which is capable of generating a collision-free path is currently being developed \cite{Cheah2021} but is beyond the scope of the work presented in this paper.
Not limited to the hardware, there is scope in improving the control system to avoid configurations which lead to the robot's wheel losing traction i.e. autonomously selecting configurations which have a CoM close to the robot's geometrical centre while adhering to other constraints such as actuator limits and obstacle avoidance.

\appendix[Control Input to Configuration Mapping Expansion]

\begin{align*}
\bm{D}_w = \bm{E}
    \begin{bmatrix}
    c_{a_1} & s_{a_1} & -l_1 s_{b_1} - l_3 c_{\alpha_1} & -l_3 c_{\alpha_1} & 0 \\
    c_{a_2} & s_{a_2} & -l_1 s_{b_2} - l_2 c_{\alpha_2} & -l_2 c_{\alpha_2} & 0 \\
    c_{a_3} & s_{a_3} & l_1 s_{b_3} - l_2 c_{\alpha_3}  & 0 & -l_2 c_{\alpha_3} \\
    c_{a_4} & s_{a_4} & l_1 s_{b_4} - l_3 c_{\alpha_4}  & 0 & -l_3 c_{\alpha_4} \\
    \end{bmatrix} 
\end{align*}
\begin{equation*}
\bm{D}_r = \frac{1}{r_r}
    \begin{bmatrix}
    s_{a_1} & -c_{a_1} & l_1 c_{b_1} + l_3 s_{\alpha_1} & -l_3 s_{\alpha_1} & 0 \\
    s_{a_2} & -c_{a_2} & l_1 c_{b_2} + l_2 s_{\alpha_2} & -l_2 s_{\alpha_2} & 0 \\
    s_{a_3} & -c_{a_3} & -l_1 c_{b_3} + l_2 s_{\alpha_3}  & 0 & -l_2 s_{\alpha_3} \\
    s_{a_4} & -c_{a_4} & -l_1 c_{b_4} + l_3 s_{\alpha_4}  & 0 & -l_3 s_{\alpha_4} \\
    \end{bmatrix}
\end{equation*}
where $\bm{E} = \text{diag}\left[r_{w_1} c_{\alpha_1}, ..., r_{w_4}c_{\alpha_4} \right]$, $a_i = \alpha_i + \phi_i + \theta$, and $b_i = \alpha_i + \phi_i$.

\section*{Acknowledgment}
The authors would like to thank Dr. Bruno Adorno for his kind support during the revision of this paper.

\ifCLASSOPTIONcaptionsoff
  \newpage
\fi

\bibliographystyle{IEEEtran}
\bibliography{IEEEabrv,library}

\begin{IEEEbiography}[{\includegraphics[width=1in,height=1.25in,clip,keepaspectratio]{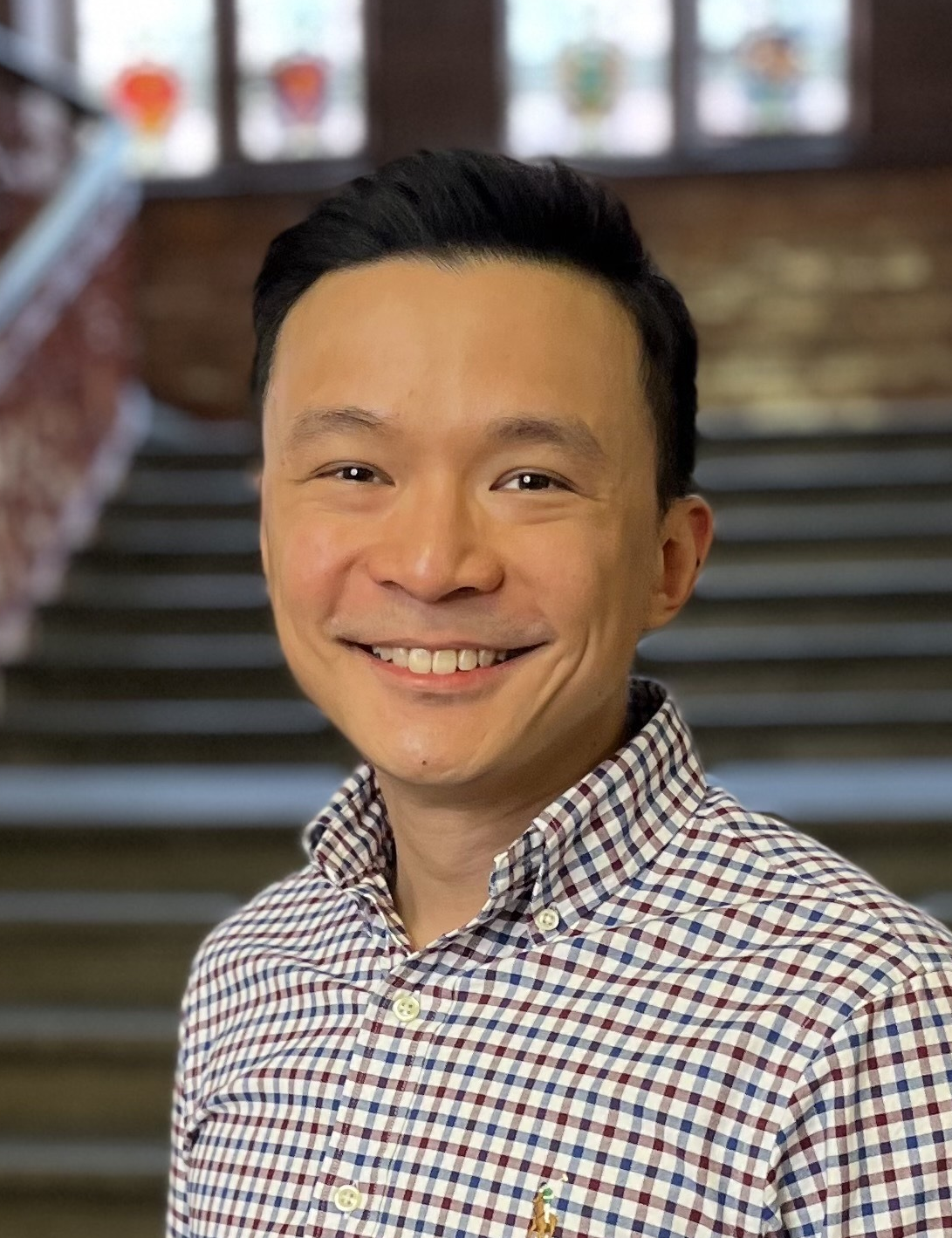}}]{Wei Cheah} received his M.Eng. degree in Mechatronic Engineering in 2015 and Ph.D. degree, both from the University of Manchester, U.K. From 2019 to 2022, he was a Post-Doctoral Research Associate with the Robotics for Extreme Environment Group within the University of Manchester. His research interests include control and motion planning for mobile robots. He is currently with Oxbotica as a software engineer developing the autonomy stack for self-driving cars.
\end{IEEEbiography}

\begin{IEEEbiography}[{\includegraphics[width=1in,height=1.25in,clip,keepaspectratio]{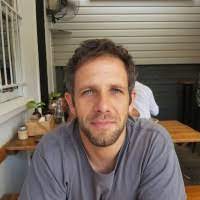}}]{Keir Groves}
is a Research Fellow at the Department of Electrical and Electronic Engineering, University of Manchester, UK. He received a first-class MEng degree in Mechanical Engineering in 2007 and a PhD in Mechanical Dynamics in 2011 from The University of Manchester. In 2017 Keir moved into Robotics and developed MallARD, the group’s autonomous aquatic surface vehicle. Keir was made Research Fellow in 2019 and specialises in localisation and control of autonomous aquatic robots for use in confined environments.
\end{IEEEbiography}
\vskip -1\baselineskip plus -1fil
\begin{IEEEbiographynophoto}{Horatio Martin}
was a research associate with the University of Manchester. His research interests are in the hardware design and control of mobile robots.
\end{IEEEbiographynophoto}
\vskip -1\baselineskip plus -1fil
\begin{IEEEbiographynophoto}{Harriet Peel}
received her MEng in Aerospace, Aeronautical and Astronautical Engineering from the University of Bristol in 2005 and a Ph.D. in Civil Engineering from University of Leeds in 2019. Her research interests are in robotics and artificial intelligence.
\end{IEEEbiographynophoto}
\vskip -1\baselineskip plus -1fil
\begin{IEEEbiography}[{\includegraphics[width=1in,height=1.25in,clip,keepaspectratio]{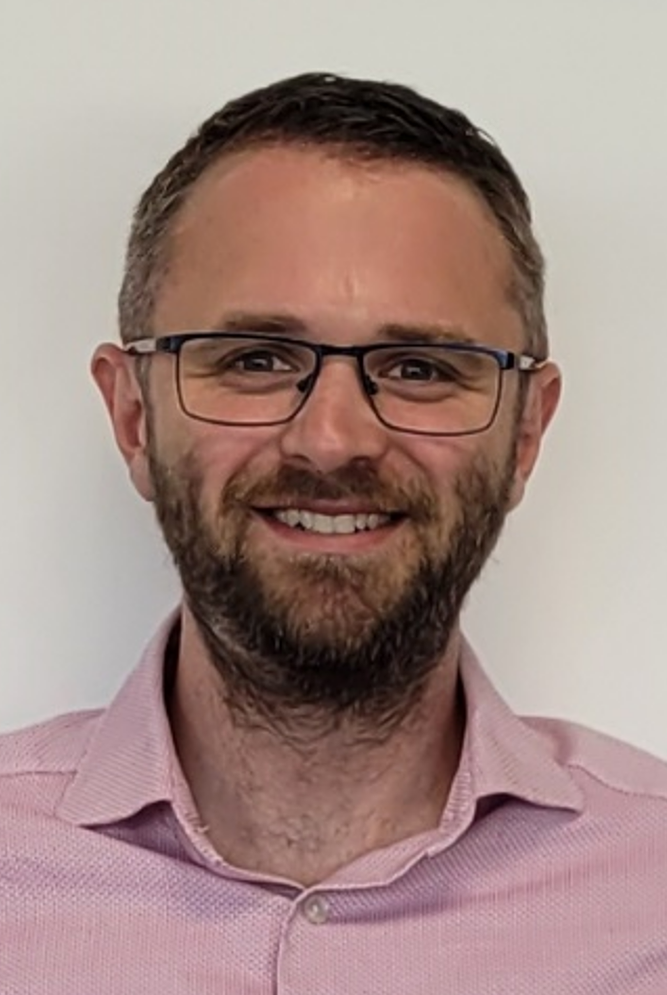}}]{Simon Watson} received his M.Eng. degree in Mechatronic Engineering in 2008 and PhD in Electrical and Electronic Engineering in 2012, both from The University of Manchester, UK.  He is currently a Reader in Robotic Systems within the Robotics for Extreme Environments Group at The University of Manchester. His research interests are focused on the development of robotic systems (aerial, aquatic and ground) for the inspection, maintenance and repair of critical energy generation infrastructure assets.
\end{IEEEbiography}

\newpage
\begin{IEEEbiography}[{\includegraphics[width=1in,height=1.25in,clip,keepaspectratio]{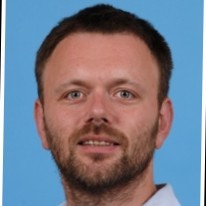}}]{Ognjen Marjanovic}
received a First Class honours degree in 1998 in Electrical and Electronic Engineering and a PhD degree in the field of Model Predictive Control in 2002 from Victoria University of Manchester. Dr Marjanovic is a Reader in Control Systems at the University of Manchester. He has over 15 years of experience working on the development and application of control and condition monitoring systems in various process industry sectors, including specialty/fine chemicals and pharmaceuticals, as well as electrical power networks and robotics.
\end{IEEEbiography}
\vskip -2 \baselineskip plus -1fil
\begin{IEEEbiography}[{\includegraphics[width=1in,height=1.25in,clip,keepaspectratio]{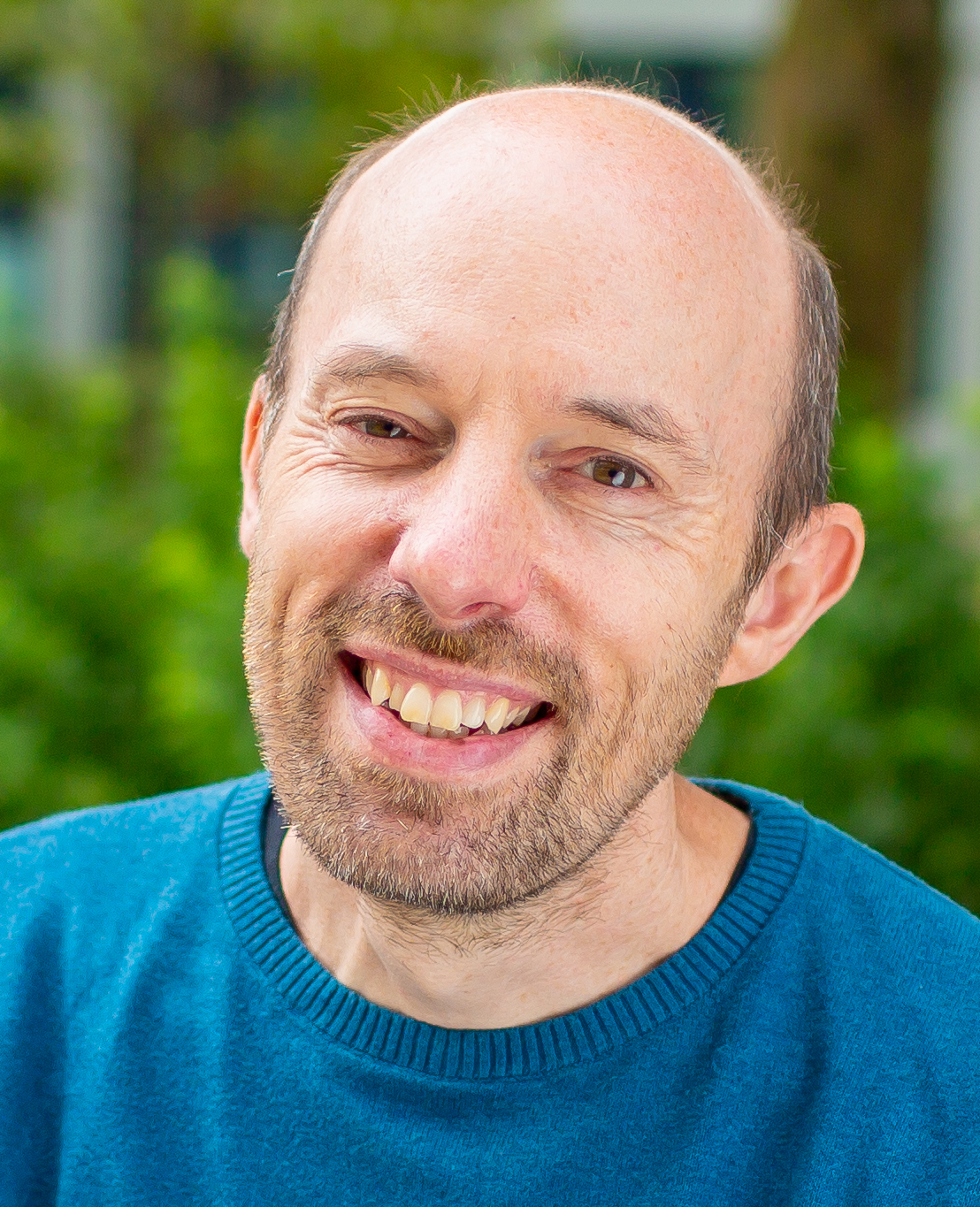}}]{Barry Lennox} received his B.Eng. degree in Chemical Engineering in 1991 and his Ph.D. in Control Systems in 1996, both from Newcastle University, UK. He is Fellow of the Royal Academy of Engineering, Professor of Applied Control and holds a Royal Academy of Engineering Chair in Emerging Technologies. His research interests are in the development of robotic systems that can used in extreme environments, with nuclear being a particular focus of his work.  
\end{IEEEbiography}





\end{document}